\documentclass[conference,10pt,a4paper]{IEEEtran}
\IEEEoverridecommandlockouts
\usepackage{cite}
\usepackage{amsmath,amssymb,amsfonts}
\usepackage{algorithmic}
\usepackage{graphicx}
\usepackage{textcomp}
\usepackage{booktabs}
\usepackage{xcolor}
\def\BibTeX{{\rm B\kern-.05em{\sc i\kern-.025em b}\kern-.08em
    T\kern-.1667em\lower.7ex\hbox{E}\kern-.125emX}}
\begin{document}

\title{Accelerating Cerebral Diagnostics with BrainFusion: A Comprehensive MRI Tumor Framework}

\author{
  \IEEEauthorblockN{%
    Walid Houmaidi\IEEEauthorrefmark{1}, 
    Youssef Sabiri\IEEEauthorrefmark{1}, 
    Salmane El Mansour Billah, 
    Amine Abouaomar
  }
  \IEEEauthorblockA{%
    School of Science and Engineering, Al Akhawayn University\\
    Ifrane, Morocco\\
    Email: \{w.houmaidi, y.sabiri, s.elmansourbillah, a.abouaomar\}@aui.ma
  }%
  \thanks{* Walid Houmaidi and Youssef Sabiri contributed equally to this work and are co‑first authors.}
}

\maketitle

\begin{abstract}
The early and accurate classification of brain tumors is crucial for guiding effective treatment strategies and improving patient outcomes. This study presents BrainFusion, a significant advancement in brain tumor analysis using magnetic resonance imaging (MRI) by combining fine-tuned convolutional neural networks (CNNs) for tumor classification —including VGG16, ResNet50, and Xception—with YOLOv8 for precise tumor localization with bounding boxes. Leveraging the “Brain Tumor MRI Dataset”, our experiments reveal that the fine-tuned VGG16 model achieves test accuracy of 99.86\%, substantially exceeding previous benchmarks. Beyond setting a new accuracy standard, the integration of bounding-box localization and explainable AI techniques further enhances both the clinical interpretability and trustworthiness of the system’s outputs. Overall, this approach underscores the transformative potential of deep learning in delivering faster, more reliable diagnoses, ultimately contributing to improved patient care and survival rates.
\end{abstract}


\section{Introduction}
Brain tumors represent a significant global health concern, with approximately 308,102 new cases and 251,329 deaths reported worldwide in 2020 \cite{globocan2020}. In the United States alone, about 24,820 malignant tumors of the brain or spinal cord are expected to be diagnosed in 2025, with an estimated 18,330 fatalities \cite{cancerstats2025}. The increasing incidence and mortality rates of primary brain tumors are partly attributed to factors such as population aging and advancements in diagnostic imaging technologies. MRI has become the gold standard for non-invasive brain tumor detection and characterization, offering superior soft-tissue contrast and high-resolution detail \cite{mri_brain}. However, the subjective nature of manual MRI interpretation poses challenges in achieving consistent and accurate diagnoses, further compounded by the complexity and heterogeneity of tumor types and the scarcity of highly specialized radiologists in many regions.

In parallel, the field of artificial intelligence (AI) has shown promising capabilities in medical image analysis, particularly through deep learning techniques \cite{deep_learning_medical}. CNNs have achieved remarkable success in various clinical tasks, including diabetic retinopathy screening, lung nodule detection, and breast cancer classification \cite{cnn_success}. Yet, the broad potential of deep learning in neuroimaging, especially for the classification of brain tumors, remains underexplored compared to other clinical domains. Given the high stakes associated with intracranial pathologies and the complexity of MRI data, there is a pressing need for robust AI-driven tools that can accurately identify, localize, and classify brain tumors to prevent clinical misdiagnosis \cite{brain_tumor_classification}.

A notable breakthrough in this context is the utilization of transfer learning and model fine-tuning. This strategy is particularly advantageous in the medical domain, where high-quality, annotated datasets are difficult to obtain due to privacy concerns, labeling costs, and data-collection complexities\cite{Gupta2019TransferLF}. Recent studies have demonstrated that fine-tuned, pre-trained CNN architectures often surpass custom-trained models, providing improved accuracy and generalizability across various classification tasks \cite{fine_tuning_cnn}.

Building on these advances, our work investigates the application of state-of-the-art transfer learning approaches for the classification of brain tumors from MRI scans. We employ three widely recognized CNN architectures—VGG16, ResNet50, and Xception—fine-tune them on the Brain Tumor MRI Dataset. 

In this study, we introduce BrainFusion, a state-of-the-art framework that provides the following pivotal contributions:

\begin{itemize}
    \item \textbf{Optimized Performance:} We fine-tune deep architectures (e.g., VGG16) to achieve outstanding accuracy, with test results exceeding 99\%, markedly improving upon traditional methods.
    \item \textbf{Streamlined Diagnostics:} Our approach balances speed and precision, equipping healthcare professionals with rapid, reliable decision-making tools.
    \item \textbf{Bounding Box Integration:} We incorporate bounding boxes to enhance the interpretability and robustness of our diagnostic framework.
    \item \textbf{Explainable AI:} We explore explainable AI techniques to bolster transparency and trust in our diagnostic models.
    \item \textbf{Enhanced Clinical Outcomes:} By addressing diagnostic delays and reducing errors, our system aims to significantly boost patient survival rates.
\end{itemize}

\section{Literature Review}
Recent advancements in AI and deep learning have significantly improved the classification of brain tumors using MRI scans. Among the most effective methods, CNNs and hybrid deep learning approaches have demonstrated state-of-the-art performance in accurately diagnosing brain tumors in the Brain Tumor MRI Dataset. Authors of \cite{Tabatabaei2023AttentionTM} proposed a novel hybrid deep learning model that integrates the Transformer Module (TM) with Self-Attention Units (SAU) and CNNs to classify brain tumors. Their model, incorporating a cross-fusion mechanism to merge local and global features, achieved an exceptional accuracy of 99.30\%, outperforming existing CNN-based methods such as iVGG (98.59\%) and iDenseNet (98.94\%) \cite{Tabatabaei2023AttentionTM}. This highlights the advantage of combining CNNs with attention-based transformer mechanisms for enhanced feature extraction in MRI-based tumor classification.

Hybrid approaches that integrate deep learning with traditional machine learning techniques have also proven highly effective. A study utilizing an ensemble approach combining low-rank Tucker decomposition with machine learning classifiers reported a high classification accuracy of 97.28\%, achieved using an extremely randomized trees model \cite{debenedictis2024tda}. Additionally, a hybrid CNN-KNN model attained 96.25\% accuracy on the BraTS dataset, outperforming standalone CNN, support vector machines (SVM), and ensemble classifiers \cite{hybrid_cnn_knn}. In another study, an SVM-based classifier with optimized feature extraction methods demonstrated an accuracy of 96.03\%, reinforcing the continued relevance of traditional machine learning approaches in medical image classification \cite{sarkar2022svm}.

Federated learning has emerged as a privacy-preserving alternative to centralized training, allowing multiple institutions to collaboratively train models without sharing sensitive patient data \cite{Das2024IndepthAO}. A federated learning framework using a modified VGG16 architecture achieved 98\% accuracy in brain tumor classification across multi-institutional MRI datasets while preserving data privacy, with precision reaching 0.99 for glioma and 0.98 for pituitary tumors \cite{federated2023mri}. This suggests that federated learning can offer competitive performance while addressing data privacy concerns.

Bounding boxes have also emerged as a pivotal tool in brain tumor detection and segmentation, offering a balance between computational efficiency and accuracy. Early methods such as the Fast Bounding Box (FBB) algorithm utilized symmetry-based approaches to detect anomalies in MRI slices, identifying regions of change by comparing intensity histograms between hemispheres. This approach enabled efficient tumor localization while avoiding challenges like intensity variations and image registration, though it required refinement for precise segmentation within the bounding box \cite{Kaur2013ThresholdingAL,Rashid2018BrainTD}. Recent advancements have incorporated deep learning, exemplified by the 3D-BoxSup framework, which leverages weakly supervised learning. By treating voxels outside the bounding box as positively labeled and those inside as unlabeled, this method mitigates inaccuracies arising from coarse annotations and achieves competitive segmentation results using minimal expert labeling. The framework demonstrated effectiveness on datasets like BraTS 2017, highlighting its potential for scalable tumor segmentation \cite{Xu20203DBoxSupPL}. Additionally, hybrid techniques combining bounding boxes with anisotropic filtering have been explored to enhance accuracy across diverse datasets, though their performance varies depending on tumor characteristics \cite{Rashid2018BrainTD}. Together, these advancements underscore the utility of bounding boxes in bridging traditional and modern segmentation techniques for brain tumor analysis.

Despite these remarkable advancements, challenges remain in ensuring model generalizability across diverse datasets and improving interpretability for clinical adoption \cite{Eche2021TowardGI}.

\section{Methods}

\subsection{Proposed System Architecture}

\begin{figure*}[!t]
    \centering
    \includegraphics[width=0.85\textwidth]{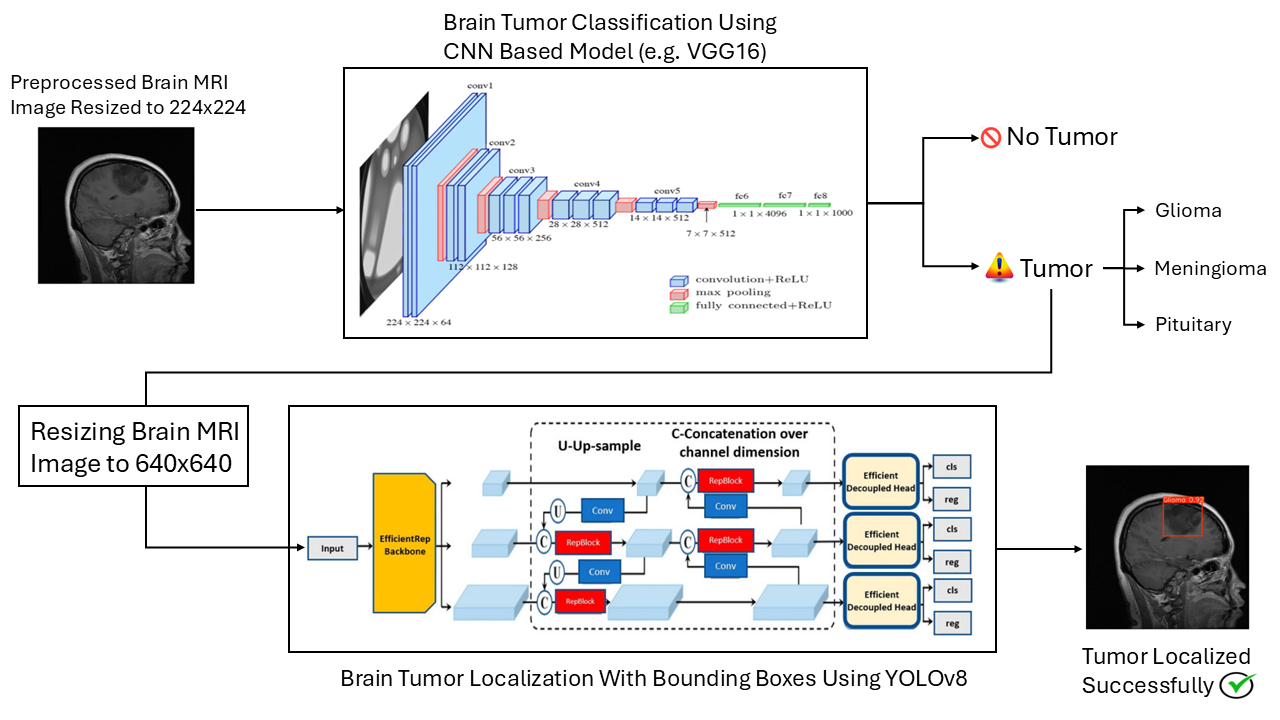}
    \caption{Proposed System Architecture}
    \label{fig:sys_archi}
\end{figure*}

Figure \ref{fig:sys_archi} illustrates BrainFusion, a two‐stage framework for classifying and localizing brain tumors in MRI scans. First, the preprocessed input brain MRI image is fed into a CNN‐based model for tumor classification, where the network determines whether a tumor is present. If the CNN detects no tumor, the process stops. However, if a tumor is identified, VGG16 further categorizes it as glioma, meningioma, or pituitary. In the second stage, the identified tumor‐positive image is passed to a YOLOv8 detector, to produce bounding boxes around the tumor region. Finally, the model outputs the MRI scan with an overlaid bounding box, clearly highlighting the localized tumor type (e.g., meningioma). This two‐phase pipeline enables robust classification and precise spatial localization of the detected brain tumor.

\subsection{Brain Tumor Classification}

\subsubsection{Dataset Description}
\begin{figure}[h]
    \centering
    \includegraphics[width=\columnwidth]{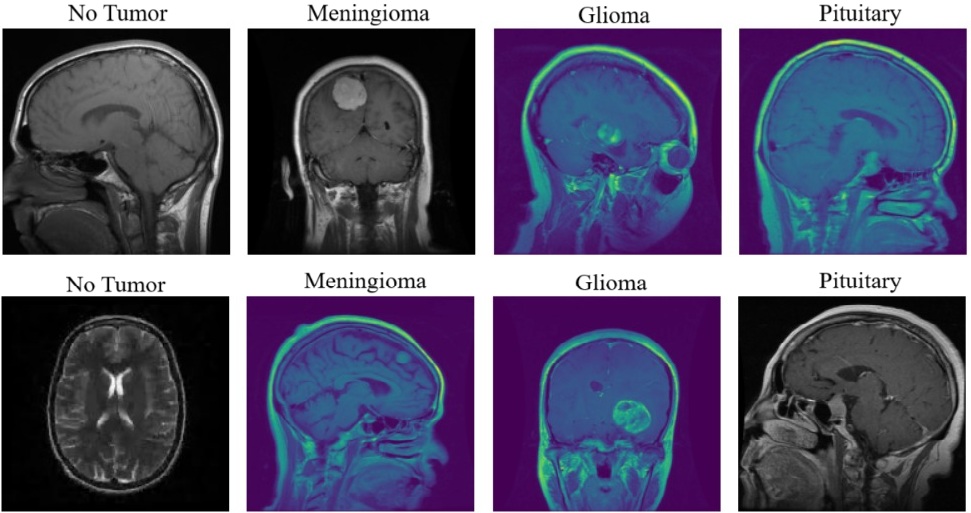}
    \caption{Sample images from the dataset, showcasing the four classes: no tumor, meningioma, glioma, and pituitary.}
    \label{fig:dataset_samples}
\end{figure}

The Brain Tumor MRI Dataset obtained from this repo \cite{nickparvar2020brain} used in this study is a merged dataset combining Figshare \cite{Cheng2017BrainTumor}, SARTAJ \cite{bhuvaji2020brain}, and Br35H \cite{hamada2020brain} datasets, containing 7,023 MRI images categorized into glioma, meningioma, pituitary tumor, and no tumor, as shown in Figure \ref{fig:dataset_samples}. The "no tumor" class is sourced from Br35H, while glioma images from SARTAJ were replaced with those from Figshare due to classification inconsistencies. The dataset offers varying image sizes, requiring preprocessing and resizing for improved model accuracy.

\subsubsection{Data Preprocessing}
Image preprocessing involved loading and resizing all MRI images to a uniform 224 × 224 pixels, ensuring consistency for CNN model input. The images were then normalized by scaling pixel values to a range of 0 to 1, enhancing model convergence and reducing computational complexity.

\subsubsection{Dataset Bias}
Despite a sizable dataset (2000 no tumor, 1645 meningioma, 1621 glioma, and 1757 pituitary images), the predominance of the no tumor class can lead to biased model performance; therefore, data augmentation is employed on the underrepresented classes to ensure balanced training and robust generalization.

\subsubsection{Dataset Splits}
The dataset was split into 80\% training, 10\% validation, and 10\% testing to ensure a well-balanced model evaluation. This structured division optimized model accuracy and generalization.

\subsubsection{Data Augmentation}
Following techniques were applied to the training data:
\begin{itemize}
    \item \textbf{Shear (30\%):} 
    Introduces subtle pixel distortions, enabling the model to learn from slight angular shifts in the data.
    
    \item \textbf{Zoom (30\%):}
    Simulates variations in distance, improving the model’s ability to handle different scales or magnifications.
    
    \item \textbf{Vertical Flip:}
    Creates mirrored images along the vertical axis, bolstering robustness against orientation changes.
    
    \item \textbf{Horizontal Flip:}
    Generates mirrored images along the horizontal axis, offering additional exposure to flipped perspectives.
    
    \item \textbf{Fill Mode (nearest):}
    Fills in any newly created pixels resulting from shear or zoom transformations with the nearest valid pixel value, preserving meaningful image content.
\end{itemize}

\subsubsection{Model Selection and Justification}
Three powerful CNN architectures—
ResNet50, VGG16, and Xception—were employed for brain tumor classification, each bringing unique strengths:
\begin{itemize}
    \item \textbf{ResNet50:} Known for its deep residual learning, it avoids vanishing gradients and captures intricate patterns. Its use of residual blocks makes it robust in capturing intricate patterns in MRI images\cite{Hofmann2022ADR}.
    \item \textbf{VGG16:} Famous for its simplicity, it can effectively capture spatial hierarchies in MRI images, making it well-suited for transfer learning, thanks to its consistent architecture with stacked convolutional layers\cite{Raghuvanshi2023TheVM}.
    \item \textbf{Xception:} An improvement over Inception networks. It leverages depthwise separable convolutions, reducing computation while enhancing performance. It excels in learning complex representations, making it suitable for the variability in tumor appearances\cite{Liu2022AnXM}.
\end{itemize}

\subsubsection{Model Architectures and Fine-Tuning}
Each model was initialized with ImageNet pre-trained weights and fine-tuned by unfreezing the last 5 layers and adding additional custom layers to adapt to the brain tumor classification task.

\subsubsection{Training Details and Hyperparameters}
The models were trained using the following hyperparameters:
\begin{itemize}
    \item \textbf{Epochs:} 100, with early stopping enabled to prevent overfitting.
    \item \textbf{Learning Rate:} Initially set to 1e-5, dynamically adjusted using the ReduceLROnPlateau scheduler.
    \item \textbf{Optimizer:} Adam, selected for its efficiency in handling sparse gradients and optimizing convergence speed.
\end{itemize}

\subsubsection{Evaluation Metrics:}

\begin{itemize}
    \item \textbf{Accuracy:} Proportion of correctly classified instances among all samples.
    \item \textbf{Precision:} Fraction of predicted positive cases that are actually positive.
    \item \textbf{Recall:} Proportion of actual positives correctly identified.
    \item \textbf{F1-Score:} Harmonic mean of Precision and Recall, balancing both metrics.
    \item \textbf{Support:} Count of true instances in each class.
\end{itemize}

Collectively, these metrics ensure a comprehensive framework for evaluating and refining classification models.
\subsection{Brain Tumor Localization Using Bounding Boxes}
\subsubsection{Dataset Description}
The dataset MRI for Brain Tumor with Bounding Boxes, obtained from this repo \cite{sorour2020mri}, comprises 5,249 meticulously annotated MRI images representing four classes of brain tumors (Glioma, Meningioma, No Tumor, and Pituitary). Each image includes a corresponding bounding box annotation in YOLO format, and covers various MRI scan angles (sagittal, axial, coronal), ensuring comprehensive anatomical representation. Derived from two existing open-source datasets and rigorously cleaned to eliminate mislabeled or low-quality samples, it provides a high-quality resource for developing and validating robust computer vision models in medical imaging.

\subsubsection{Data Preprocessing}
All images and labels are consolidated within a unified directory structure compatible with YOLO-based training. Ground-truth annotations were converted into YOLO's normalized format, specifying the class, \(x_{\text{center}}\), \(y_{\text{center}}\), width, and height relative to each image.

\subsubsection{Dataset Splits}
The dataset was split into 80\% training, 10\% validation, and 10\% testing.

\subsubsection{Data Augmentation}
During training, Albumentations is employed to perform automated transformations—including random flips, scaling, color jitter, and slight blur—to simulate real-world variations in tumor appearance. These augmentations prevent overfitting and enhance the model's generalization.

\subsubsection{Model Selection and Justification}
YOLOv8 achieves state-of-the-art detection accuracy while maintaining low-latency inference, making it ideal for real-time or large-scale medical applications \cite{Widayani2024ReviewOA}.

\subsubsection{Model Fine-Tuning}
The pretrained YOLOv8 model is adapted to the specific brain tumor dataset by fine-tuning all its layers, retaining valuable feature representations while refining the detection layer for new classes.

\subsubsection{Training Details and Hyperparameters}
\begin{itemize}
    \item \textbf{Epochs:} Set to 30 epochs
    \item \textbf{Batch Size:} Set to 16.
    \item \textbf{Input Resolution:} Images are resized to $640 \times 640$, a standard for YOLO-based models that preserves sufficient detail while managing memory constraints.
    \item \textbf{Optimizer and LR Strategy:} An AdamW optimizer with automated built-in learning rate scheduling ensures dynamic adjustment during training.

\end{itemize}

\subsubsection{Evaluation Metrics}
We used not only \textbf{Precision} and \textbf{Recall} but also evaluation metrics such as \textbf{mAP@0.5} and \textbf{mAP@0.5--0.95} to comprehensively assess model performance. Specifically, mAP@0.5 is calculated as the average of class-wise average precisions at an IoU threshold of 0.5, while mAP@0.5--0.95 extends this evaluation by averaging the class-wise average precisions over IoU thresholds ranging from 0.5 to 0.95.

\section{Results and Discussion}
\subsection{Brain Tumor Classification}

\subsubsection{Performance Comparison}
\begin{table}[ht]
\centering
\setlength{\tabcolsep}{4pt}
\renewcommand{\arraystretch}{1.2}
\caption{Performance Comparison on the Brain Tumor MRI Dataset}
\label{tab:performance_comparison}
\begin{tabular}{c|l|c|l}
\hline
\textbf{Rank} & \textbf{Method/Study} & \textbf{Accuracy} & \textbf{Reference} \\
\hline
1 & \textbf{Ours: Fine-Tuned VGG16}          & \textbf{99.86\%} & -- \\
2 & Hybrid DL (Transformer + SAU + CNN)      & 99.30\% & \cite{Tabatabaei2023AttentionTM} \\
3 & iDenseNet                                & 98.94\% & \cite{Tabatabaei2023AttentionTM} \\
4 & iVGG                                     & 98.59\% & \cite{Tabatabaei2023AttentionTM} \\
5 & SVM-based Classifier                     & 98.30\% & \cite{sarkar2022svm} \\
6 & \textbf{Ours: Fine-Tuned ResNet50}       & \textbf{98.15\%} & -- \\
7 & Federated Learning (VGG16)                 & 98\% & \cite{federated2023mri}  \\
8 & \textbf{Ours: Fine-Tuned Xception}       & \textbf{97.41\%} & -- \\
9 & Tucker Decomposition + ML Classifiers    & 97.28\% & \cite{debenedictis2024tda} \\
10 & Hybrid CNN-KNN                          & 96.25\% & \cite{hybrid_cnn_knn} \\
\hline
\end{tabular}
\end{table}

In Table~\ref{tab:performance_comparison}, we compare our results against notable benchmarks in the literature. Our fine-tuned VGG16 model achieves \textbf{the highest accuracy (99.86\%)}, surpassing even recent transformer-based and advanced CNN methods. Our experiments underscore the potential of carefully fine-tuned deep architectures to establish new state-of-the-art results on the Brain Tumor MRI Dataset.
\subsubsection{Classification Report Analysis}

\begin{table}[h]
\centering
\setlength{\tabcolsep}{6pt}
\renewcommand{\arraystretch}{1}
\caption{Classification Report -- VGG16 Model Performance Metrics}
\label{tab:classification-metrics}
\begin{tabular}{lcccc}
\toprule
\textbf{Class} & \textbf{Precision} & \textbf{Recall} & \textbf{F1-score} & \textbf{Support} \\
\midrule
Glioma      & 1.00 & 0.99 & 0.99 & 300 \\
Meningioma  & 0.99 & 0.99 & 0.99 & 306 \\
No Tumor    & 1.00 & 1.00 & 1.00 & 405 \\
Pituitary   & 0.99 & 1.00 & 1.00 & 300 \\
\midrule
\textbf{Accuracy}   & \multicolumn{3}{c}{1.00} & 1311 \\
\textbf{Macro Avg}  & 1.00 & 1.00 & 1.00 & 1311 \\
\textbf{Weighted Avg} & 1.00 & 1.00 & 1.00 & 1311 \\
\bottomrule
\end{tabular}
\end{table}
The classification report (Table \ref{tab:classification-metrics}) shows that the model performs exceptionally well across all tumor classes, with high precision, recall, and F1-score values. The near-perfect accuracy (1.00) and balanced macro and weighted averages indicate that the VGG16 model generalizes effectively across different tumor categories. Notably, the model achieved perfect classification for No Tumor and Pituitary Tumor, underscoring its reliability and robustness in distinguishing between tumor and non-tumor MRI images.
\begin{figure*}[!t]
    \centering
    \includegraphics[width=0.75\textwidth]{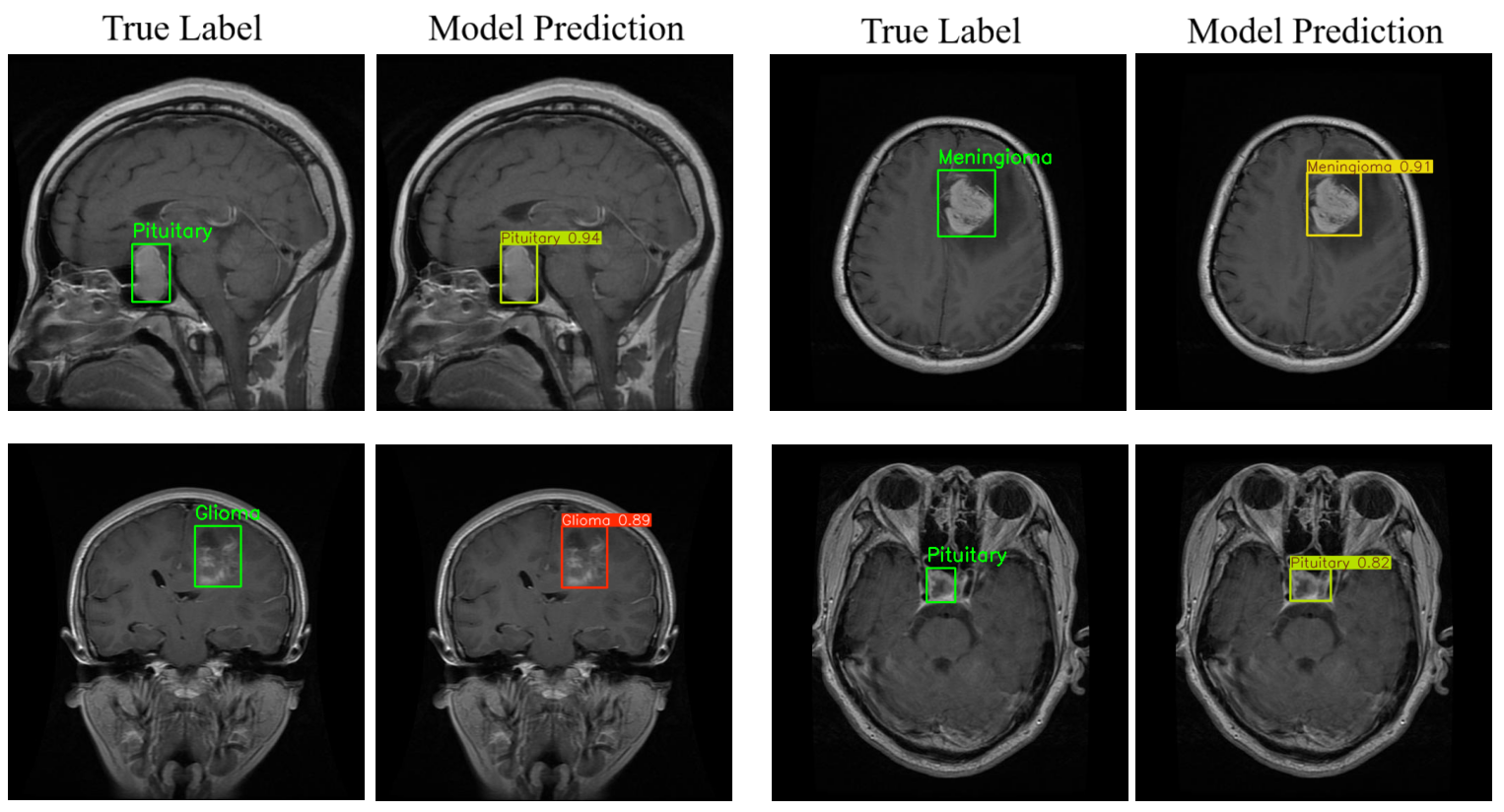}
    \caption{True Labels and YOLOv8 Model Predictions for Brain Tumor Localization}
    \label{fig:boxes_brain}
\end{figure*}

\begin{figure*}[!t]
    \centering
    \includegraphics[width=0.75\textwidth]{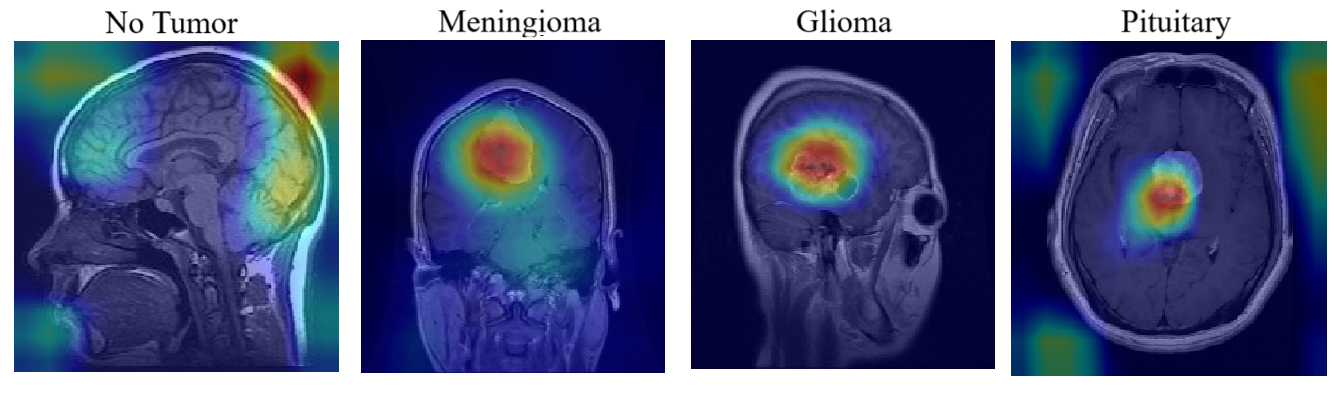}
    \caption{Grad-CAM Visualizations for Brain Tumor MRI Classification}
    \label{fig:gradcam}
\end{figure*}

\subsubsection{Interpretation of Results}
VGG16's deep, sequential convolutional layers enable superior feature extraction for brain tumor classification, outperforming ResNet50 and Xception; minor misclassifications between glioma and meningioma reflect intrinsic tumor similarities rather than model inefficiency, underscoring the promise of transfer learning for robust MRI-based diagnostics.

\subsection{Brain Tumor Localization Using Bounding Boxes}

\begin{table}[h]
\centering
\setlength{\tabcolsep}{1.5pt} 
\renewcommand{\arraystretch}{1} 
\caption{YOLOv8 Detection Results Summary on the Test Set}
\label{tab:detection_summary}
\begin{tabular}{lcccccc}
\toprule
\textbf{Class} & \textbf{Images} & \textbf{Instances} & \textbf{Precision} & \textbf{Recall} & \textbf{mAP@0.5} & \textbf{mAP@0.5--0.95} \\
\midrule
Glioma     & 132 & 152 & 0.881 & 0.825 & 0.910 & 0.701 \\
Meningioma & 167 & 172 & 0.955 & 0.983 & 0.986 & 0.834 \\
No Tumor   &  75 &  75 & 0.994 & 0.987 & 0.995 & 0.839 \\
Pituitary  & 181 & 191 & 0.925 & 0.900 & 0.956 & 0.749 \\
\midrule
\textbf{All} & 526 & 590 & 0.939 & 0.924 & 0.962 & 0.781 \\
\bottomrule
\end{tabular}
\end{table}

The YOLOv8 model demonstrates high detection accuracy across all tumor categories, as indicated by both strong \textbf{Precision} (0.939) and \textbf{Recall} (0.924) in Table \ref{tab:detection_summary}. This balance suggests that the model not only correctly localize most tumor instances (high recall), but also makes few incorrect predictions (high precision). Furthermore, the high \textbf{mAP@0.5} (0.962) indicates that the predicted bounding boxes overlap well with ground-truth annotations, while the \textbf{mAP@0.5--0.95} (0.781) evidences robust performance at more stringent overlap thresholds. Among the tumor classes, ``Meningioma'' attains the highest recall (0.983), indicating very few missed tumors. Overall, these metrics confirm the model's effectiveness in consistently localizing various brain tumor types.

\section{Explainable AI (XAI) for Brain Tumor Detection}
Modern deep learning models often operate as ``black boxes'', making it challenging to interpret how they arrive at certain conclusions. In a medical context---especially concerning life-critical decisions like tumor detection---transparency is paramount. XAI bridges this gap by revealing the model's decision-making process. Here, bounding boxes as shown in Figure \ref{fig:boxes_brain} provide localized evidence of detected tumors, indicating the exact regions of interest, while Grad-CAM (Gradient-weighted Class Activation Mapping) \cite{gradcam} as shown in Figure \ref{fig:gradcam} visually highlights the most influential features within an image. These tools help clinicians understand why the model flags certain areas as problematic, thus instilling confidence in its outputs. Moreover, explainability is crucial for regulatory compliance, ethical considerations, and patient trust, as it allows practitioners to validate automated findings and ensure that misclassifications can be quickly identified and addressed. By combining high accuracy with interpretability, our approach supports safer, more accountable diagnostic workflows in brain tumor detection.

\section{Conclusion and Future Work}
This paper presented BrainFusion, a comprehensive MRI tumor framework that integrates transfer learning for brain tumor classification and bounding-box localization for precise tumor detection. By fine-tuning state-of-the-art CNN architectures—VGG16, ResNet50, and Xception—we demonstrated that our approach, particularly the fine-tuned VGG16 achieving 99.86\% accuracy, surpasses existing methods in both classification accuracy and diagnostic reliability. Moreover, the incorporation of YOLOv8 for tumor localization and Grad-CAM for visual explanations enhances the system's clinical interpretability and trustworthiness. Our findings underscore the transformative potential of deep learning for rapid, automated, and explainable brain tumor diagnosis, while highlighting the critical role of robust data augmentation and preprocessing in medical image analysis.

Despite these promising results, several challenges remain. Future research should investigate domain adaptation techniques to improve generalizability across diverse MRI protocols, further refine explainable AI methods to clarify model decision processes, and explore federated learning to facilitate collaborative training across institutions while preserving data privacy. Addressing these issues will enhance the clinical utility of BrainFusion and similar AI-driven diagnostic tools.

\bibliographystyle{IEEEtran}
\bibliography{references}

\end{document}